\documentclass[sigconf]{acmart}
 
\usepackage[utf8]{inputenc}
\usepackage[T1]{fontenc}
\usepackage{soul}
\usepackage{balance} 

\usepackage{mathtools,xparse}

\DeclarePairedDelimiter{\norm}{\lVert}{\rVert}
\usepackage{arydshln}

\def\BibTeX{{\rm B\kern-.05em{\sc i\kern-.025em b}\kern-.08emT\kern-.1667em\lower.7ex\hbox{E}\kern-.125emX}}

\copyrightyear{2019}
\acmYear{2019}
\acmConference[MM '19]{Proceedings of the 27th ACM International Conference on
Multimedia}{October 21--25, 2019}{Nice, France}
\acmBooktitle{Proceedings of the 27th ACM International Conference on Multimedia (MM '19),
October 21--25, 2019, Nice, France}
\acmPrice{15.00}
\acmDOI{10.1145/3343031.3350908}
\acmISBN{978-1-4503-6889-6/19/10}

\usepackage{arydshln}

\begin{document}
\fancyhead{}

\title{IntersectGAN: Learning Domain Intersection for Generating Images with Multiple Attributes}

\author{Zehui Yao$^{1}$, Boyan Zhang$^{1}$, Zhiyong Wang$^{1}$, Wanli Ouyang$^{2,3}$, Dong Xu$^{2}$, Dagan Feng$^{1}$}
\affiliation{
    \institution{$^{1}$School of Computer Science, The University of Sydney, Australia}
     \institution{$^{2}$School of Electrical and Information Engineering, The University of Sydney, Australia}
    \institution{$^{3}$The University of Sydney, SenseTime Computer Vision Research Group, Australia}
}
\email{{zyao3112,bzha8220}@uni.sydney.edu.au, {zhiyong.wang,wanli.ouyang,dong.xu,dagan.feng}@sydney.edu.au}

\begin{abstract}

Generative adversarial networks (GANs) have demonstrated great success in generating various visual content.
However, images generated by existing GANs are often of attributes (e.g., smiling expression) learned from one image domain.
As a result, generating images of multiple attributes requires many real samples possessing multiple attributes which are very resource expensive to be collected.
In this paper, we propose a novel GAN, namely IntersectGAN, to learn multiple attributes from different image domains through an intersecting architecture.
For example, given two image domains $X_1$ and $X_2$ with certain attributes, the intersection $X_1 \cap X_2$ denotes a new domain where images possess the attributes from both $X_1$ and $X_2$ domains.
The proposed IntersectGAN consists of two discriminators $D_1$ and $D_2$ to distinguish between generated and real samples of different domains, and three generators where the intersection generator is trained against both discriminators.
And an overall adversarial loss function is defined over three generators. 
As a result, our proposed IntersectGAN can be trained on multiple domains of which each presents one specific attribute, and eventually eliminates the need of real sample images simultaneously possessing multiple attributes.
By using the CelebFaces Attributes dataset, our proposed IntersectGAN is able to produce high quality face images possessing multiple attributes (e.g., a face with black hair and a smiling expression).
Both qualitative and quantitative evaluations are conducted to compare our proposed IntersectGAN with other baseline methods. Besides, several different applications of IntersectGAN have been explored with promising results.
\end{abstract}


%

\maketitle

\begin{figure}[h!]
\begin{center}

\includegraphics[height=55mm]{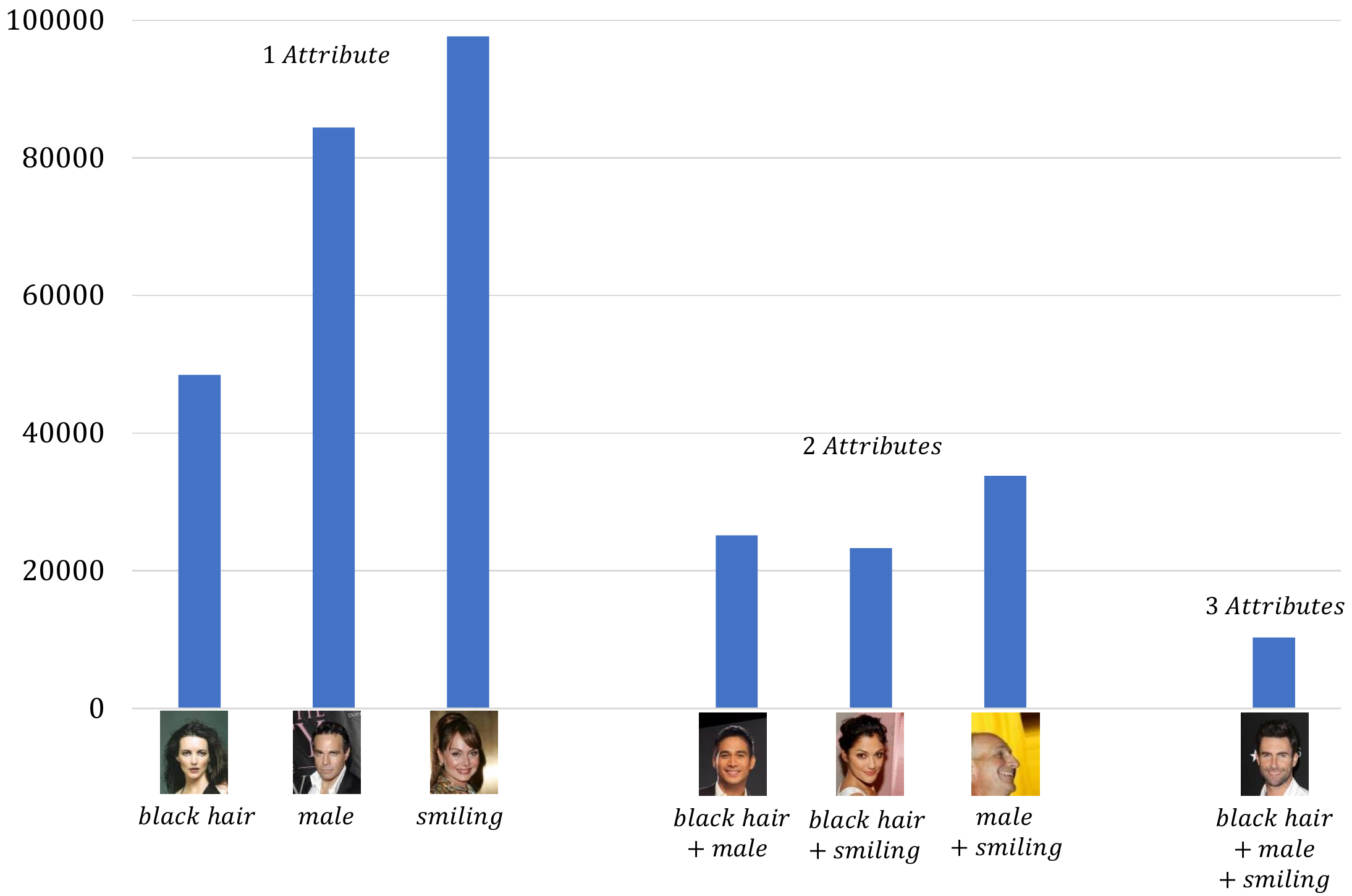}
\end{center}
  \caption{Statistics of the images with single attribute and multiple attributes in the CelebFaces Attributes dataset. The sample attributes and images are for illustration purpose.
  }
\label{fig:attribute_chance}
\end{figure}


\section{Introduction}
\label{sec:intro}


In recent years, generative adversarial networks (GANs) have achie- ved promising progress on image generation due to its adversarial training strategy. 
A GAN model generally consists of two key components, a generator and a discriminator. The generator outputs generated samples (i.e., fake samples) that are as indistinguishable as possible from real samples, while the discriminator differentiates real samples from fake samples (i.e., generated samples) as much as possible. 
As a result, the generator is able to match the distribution of real samples.

Motivated by the success of generating synthetic data (e.g., digits \cite{GAN}) for training a better classifier, various GANs-based frameworks have been proposed for different applications such as image style transfer \cite{Pix2Pix, ZhangStyle, YangStyle},   image super resolution \cite{SRGAN},  high resolution image generation \cite{PGGAN, HDPix2pix},  text-based image generation \cite{StackGAN, AttnGAN},
faces image synthesis \cite{Face2, Face3, StarGAN} and fashion designs \cite{GAN_fashion, PoseImageGenration}. 

When generating images with multiple attributes using conventional GANs,  we often need to collect real samples possessing all the required attributes and train the networks with the collected images or apply an extension of GAN to a conditional setting.
However, as shown in Figure \ref{fig:attribute_chance}, when the number of specified attributes increases, in general, the number of samples decrease dramatically. It is challenging to collect a large number of real samples possessing multiple attributes for training deep networks, and also resource expensive to produce supervision labels.
It is anticipated that it will be increasingly challenging when the number of attributes increases.



\begin{figure}[t]
\begin{center}
\includegraphics[height=53mm]{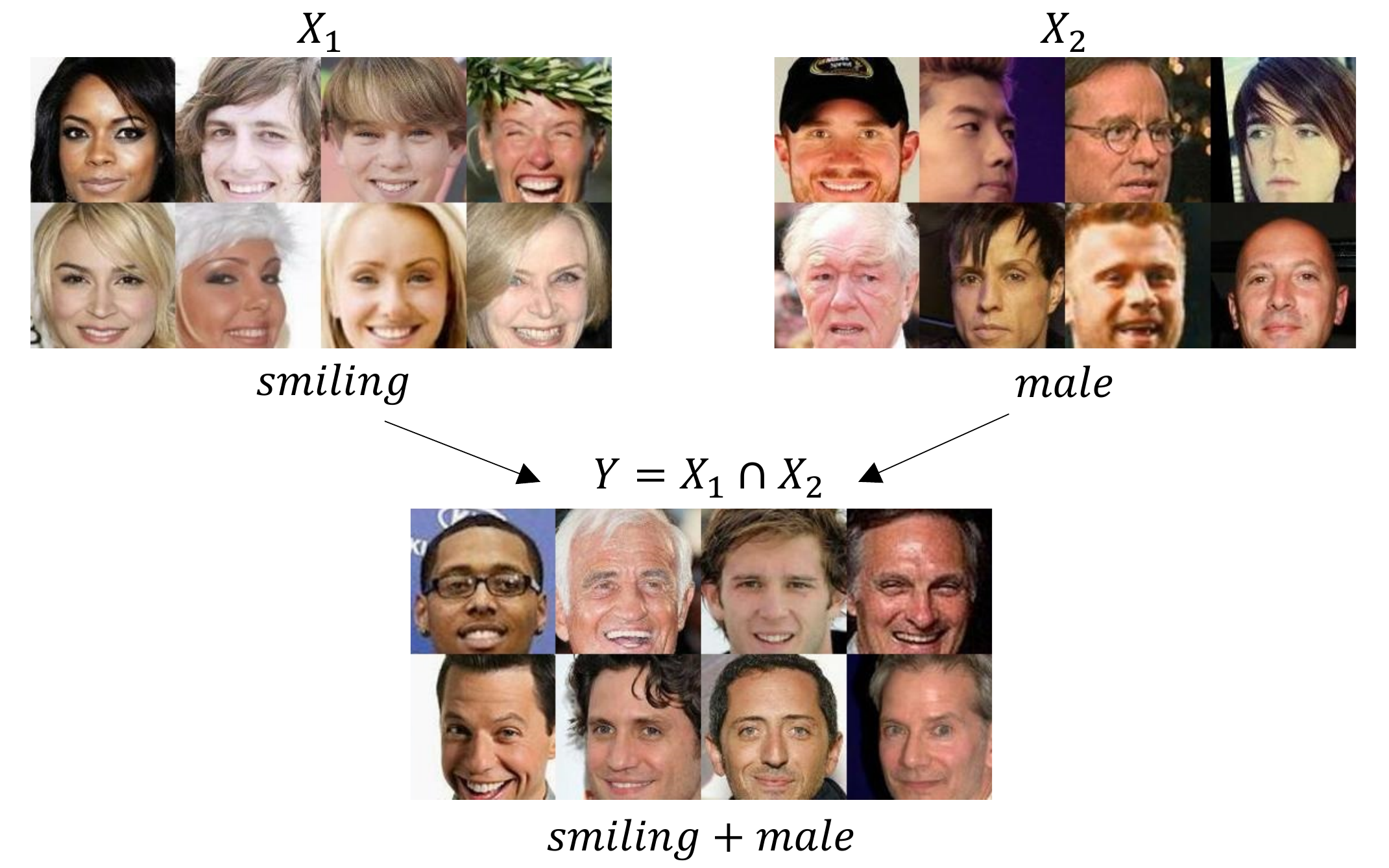}
\end{center}
  \caption{Illustration of intersecting two image domains $X_1$ and $X_2$ into a new domain $X_1 \cap X_2$. 
  Domain $X_1$ contains face samples with attribute \textit{smiling}, domain $X_2$ contains face samples with attribute \textit{male}, and domain $X_1 \cap X_2$ represents face images with both \textit{smiling} and \textit{male}.}
\label{fig:data_example}
\end{figure}

To overcome such limitation of the existing GANs, we propose a novel GAN model, namely IntersectGAN, to generate images with multiple attributes by learning the distribution of a new domain intersected from existing domains.
For simplicity of explanation, we use two domains in the following discussions.
Given two image domains $X_1$ and $X_2$ with corresponding attributes ${{{A}}}_1$ and ${{{A}}}_2$, respectively. 
The intersection $Y = X_1 \cap X_2$ denotes a new domain where images possess the attributes from both $X_1$ and $X_2$ domains, and the new intersected attribute is denoted as ${{{A}}}_{Y}$.
That is, the intersected domain $X_1 \cap X_2$ possesses a new attribute ${{{A}}}_{Y}$ denoted as ${{{A}}}_1 \cup {{{A}}}_2$, which is the combination of both attributes ${{{A}}}_1$ and ${{{A}}}_2$. 
As illustrated in Figure \ref{fig:data_example}, ${{{A}}}_1$ is \textit{smiling} and ${{{A}}}_2$ is \textit{male}. The images in the intersected domain $X_1 \cap X_2$ possess these two attributes simultaneously. That is, ${{{A}}}_{X_1 \cap X_2} = {{{A}}}_1 \cup {{{A}}}_2 = \{\textit{smiling}, \textit{male} \}$.

\begin{figure*}
\begin{center}
\includegraphics[height=70mm]{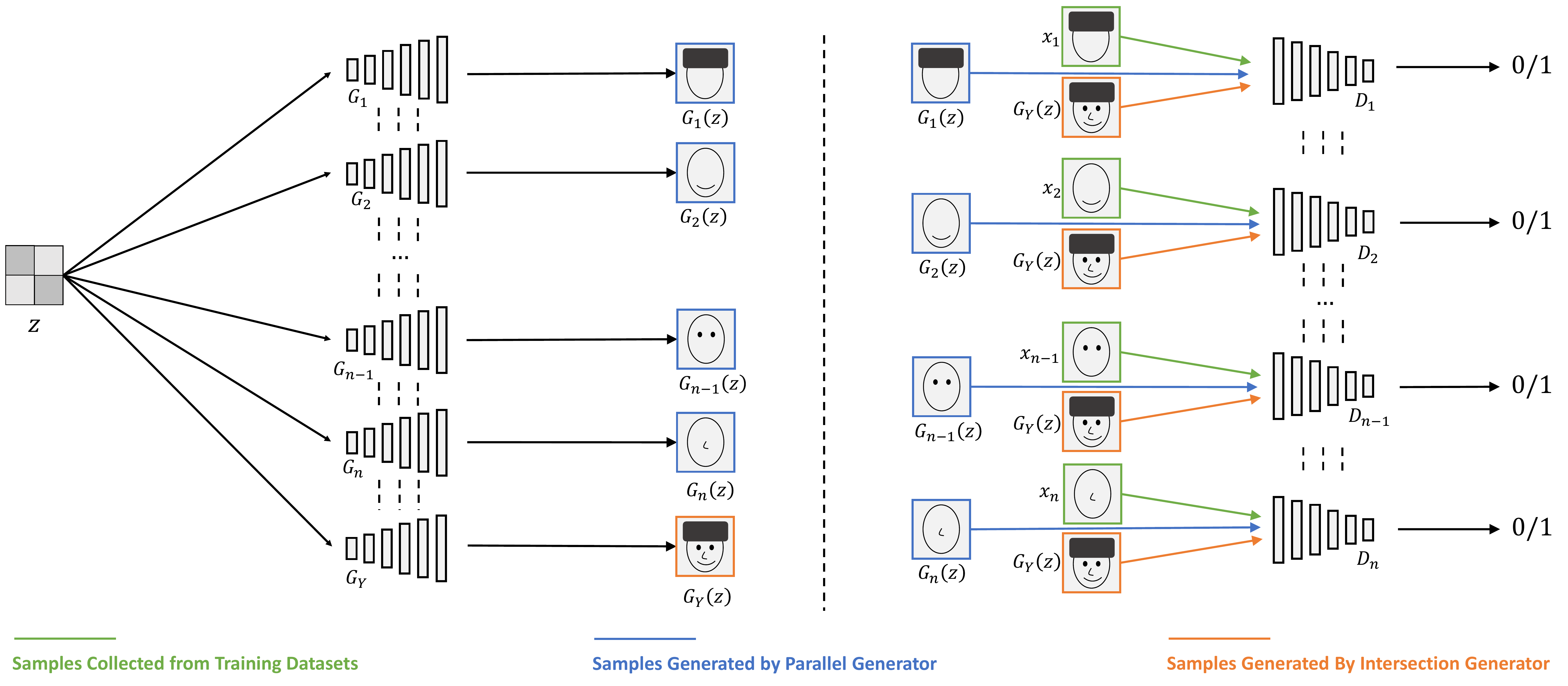}
\end{center}
  \caption{Illustration of our proposed IntersectGAN model which consists of $n$ parallel generators $G_1$, $G_2$, ... $G_{n-1}$, $G_n$ and an intersection generator $G_{Y}$ as well as $n$ discriminators $D_1$, $D_2$, ... $D_{n-1}$ and $D_n$. 
  Note that $z$ denotes the random noise input, $x_1$, $x_2$, ... $x_{n-1}$ and $x_n$ are the real samples collected from the corresponding image domains $X_1$, $X_2$, ... $X_{n-1}$ and $X_n$, respectively. 
  The generators produce fake image samples $G_1(z)$, $G_2(z)$, ..., $G_{n-1}(z)$, $G_n(z)$ and $G_{Y}(z)$, while the discriminators would output a value ranging from 0 to 1 indicating the probability whether an input image is a real sample. 
  The dash lines between generators (discriminators) represent optional modifications on the network architecture (e.g., weight sharing).
  }
\label{fig:IntersectGAN_model}
\end{figure*}

As illustrated in Figure \ref{fig:IntersectGAN_model}, when generating images with $n$ specified attributes, our proposed IntersectGAN model consists of $n+1$ generators (i.e., one generator for each attribute and one generator for combined attributes) and $n$ discriminators (i.e., one for each attribute). 
In comparison with conventional GANs, the discriminators are further challenged to be more discriminant by also taking the input from the intersection generator $G_{Y}$.
During the training phase, the intersection generator aims to produce content which is as indistinguishable as possible to all $n$ image domains.
We define the overall adversarial loss as the summation of the losses from $n+1$ generators.
That is, the intersection generator $G_{Y}$ is trained to fit the distribution of samples possessing the attributes from all the given domains.
As a result, our proposed IntersectGAN is able to produce synthesized samples possessing multiple attributes without relying on real samples simultaneously possessing those attributes.

Furthermore, we explore the capacity of our proposed IntersectGAN in several other image generation tasks, such as generating images of a blended attribute by intersecting two opposite attributes (e.g., {\it male} and {\it female}) and content-aware domain intersection.
In addition, by exploiting the interaction among three generators with inter-layer weight sharing, we are able to adapt the IntersectGAN to generate trio image samples, which better illustrates the idea of learning domain intersection.

In summary, the key contributions of our work are as follows:
\begin{enumerate}

\item We propose a novel IntersectGAN model which is able to generate image samples possessing multiple attributes without using real samples simultaneously possessing those attributes. 
To the best of our knowledge, this is the first GAN model specially developed for generating images with multiple attributes from noise input by learning a domain intersection without introducing supervision labels.

\item We formulate the intersection learning problem under the GAN framework to learn the distribution of an intersection of multiple domains. 
New adversarial training strategies are proposed to take into account the new architecture: the discriminators are trained against the parallel generators of individual domains plus the intersection generator while the intersection generation is trained against all the discriminators. 

\item We demonstrate the capacity of the proposed IntersectGAN in other image generation applications, such as generating images with a blended attribute and generating content-aware domain intersected images.
In particular, a weight sharing architecture (as illustrated in dashed lines in Figure \ref{fig:IntersectGAN_model}) is proposed to explore the relationship among generators and generate trio images.
More details of weight sharing are explained in Section \ref{ssec:trio}.

\end{enumerate}

\section{Related Work}
\label{sec:review}

In this section, we organize GAN based image generation methods into two categories: image generation and image-to-image translation, in terms of whether a real input image is needed for the generator.
The first category aims to generate image samples from a noise vector after learning the distribution of an image domain, while the second category aims to generate a new version of each given image by translating the input sample to another image domain.

\subsection{Image Generation}

GAN \cite{GAN} was first proposed to learn a target distribution with an adversarial learning strategy, and has dramatically accelerated the progress on image generation. 
It has been widely utilized to generate image samples of the same style as of the given training image set.
Various other improvements for GANs have been developed, such as network architectures, loss functions, and optimization strategies.
For example, DCGAN \cite{CDGAN} was proposed to boost the generative performance of GAN with a new architecture consisting of convolution layers instead of using max pooling or fully connected layers.
Many GANs have also been proposed by using new forms of cost functions (e.g., L2 loss and L1 loss etc.) to boost the training process, such as LSGAN \cite{LSGAN}, WGAN \cite{WGAN}, WGAN-GP \cite{WGAN_GP}.
However, these GANs take only one image set (i.e., image domain) and follow the one-to-one adversarial learning strategy, namely, one generator is trained against one discriminator.
When generating images of multiple attributes through these GANs, real samples simultaneously possessing those attributes are required. However, building qualified dataset is expensive and sometimes even impractical.

In order to influence the content of a generated image in addition to the style, conditional GAN (cGAN) \cite{ConditionalGAN} was proposed by adding content labels as an additional input to both generator and discriminator.
As a result, the content of generated images is supposed to be of the same as of the specified label, and the style of generated images will be of the same as of the given training images.  
However, as a supervised learning approach, even for its state-of-the-art cGAN with Projection Discriminator \cite{cgan-pd}, cGAN requires a large number of samples with supervision labels, which are expensive to obtain.
Our proposed IntersectGAN is able to address this limitation through its intersection architecture.
That is, only the samples possessing individual attributes are required to train IntersectGAN for generating images with multiple attributes and no supervision labels are systematically needed.

\subsection{Image-to-Image Translation}

Image-to-image translation is a vision and graphics problem aiming to learn a mapping between an input image domain and a target image domain with two training datasets (one dataset for each domain).
Pix2pix \cite{Pix2Pix} was first proposed to perform one-way image-to-image translation through a supervised learning approach.
A pair of samples (e.g., a sketch image and a color image) is required for both the generator and discriminator.
In order to loose the requirement on paired training samples, various GANs have been proposed, such as UNIT \cite{UNIT} and CycleGAN \cite{CycleGAN}, DiscoGAN \cite{DiscoGAN}, DualGAN \cite{DualGAN}.
However, these GANs are limited to translate a given image of one domain to an image of the other domain.

Three recently proposed GANs, IcGAN \cite{IcGAN}, conditional CycleGAN \cite{conditionalCycleGAN} and StarGAN \cite{StarGAN}, are able to translate a given image into an image with multiple attributes through conditional input labels.
However, label information is required during the training as the conditional configuration. 
Differently, our IntersectGAN is proposed for image generation from noise input, instead of image-to-image translation, and supervision label information is not required during training.

\section{IntersectGAN}
\label{sec:method}

In this section, we first present the formulation of IntersectGAN and then explain its implementation details.

\subsection{Formulation}
Given image domains $X_1$, $X_2$ ... $X_n$ specified with $n$ attributes ${{{A}}}_1$, ${{{A}}}_2$ ... ${{{A}}}_n$, respectively, IntersectGAN aims to learn a mapping from random noise to the target domain $Y = \bigcap_{1 \le i \le n}{X_i}$ where generated samples are supposed to possess all the specified attributes. 
We denote the distribution of real image samples as $x_i \sim p_{X_{i}}$ for $1 \le i \le n$. 
In addition, the distribution of random noise is denoted as $z \sim p_Z$. 
During the training phase, our model learns $n$ \textit{parallel mappings} $G_i: Z \rightarrow X_{i}$ and an \textit{intersection mapping} $G_{\mathrm{Intersect}}: Z \rightarrow Y$. 
Meanwhile, there are $n$ discriminators $D_{1}$, $D_{2}$ ... $D_{n}$ distinguishing between real and generated samples with regard to the corresponding domains $X_{1}$, $X_{2}$ ... $X_{n}$, respectively. 
In the training stage, the expected output of a discriminator for a real image is set to 1, and that for a generated image is set to 0.

Motivated by GAN, our proposed IntersectGAN is formulated through the definition of adversarial loss.
According to the objective of GAN, for an image domain $X$, a generator $G$ and a corresponding discriminator $D$, the adversarial loss function is defined as:

\begin{equation}
\begin{split}
\mathcal{L}_{\mathrm{GAN}}(X, G, D) & = \  \mathbb{E}_{x \sim p_X} \left[ \log{D(x)} \right] \\
& + \ \mathbb{E}_{z \sim p_Z} \left[ \log{(1 - D(G(z)))} \right],
\end{split}
\label{eqt: gan_loss}
\end{equation}
where $G$ is trained to generate synthetic images $G(z)$ possessing the attribute of domain $X$ while $D$ tries to differentiate real image samples $x$ from generated samples $G(z)$.

As illustrated in Figure \ref{fig:IntersectGAN_model}, IntersectGAN has $n$ \textit{parallel generators} $G_1$, $G_2$, ... $G_n$ as well as $n$ corresponding discriminators $D_1$, $D_2$, ..., $D_n$. 
Their adversarial losses can be expressed as:
\begin{equation}
\mathcal{L}_{i} = \alpha_i \mathcal{L}_{\mathrm{GAN}}(X_i, G_i, D_i), \ \mathrm{for} \ 1 \le i \le n,
\label{eq: parallel_gan_loss}
\end{equation} 
where $\alpha_i$ indicates the objective importance of the $i$-th parallel generator.

The intersection generator $G_{Y}$ produces image samples containing all the specified attributes ${{{A}}}_1$, ${{{A}}}_2$ ... ${{{A}}}_n$. Ideally, each discriminator $D_i$ cannot distinguish between the generated samples $G_{Y}(z)$ and real images from its corresponding domain $X_i$.  
Therefore, we can formulate the objective of $G_{Y}$ as a weighted sum of $n$ adversarial losses:
\begin{equation}
\mathcal{L}_{\mathrm{Intersect}} = \ \sum_{1 \leq i \leq n}{ \alpha_i \mathcal{L}_{\mathrm{GAN}}(X_i, G_{Y}, D_{i}),} 
\label{eq: intersect_gan_loss}
\end{equation}
where $\alpha_i$ is the importance constant mentioned above and $G_{Y}$ aims to simultaneously minimize two adversarial losses against both discriminators. 

For the whole model, the overall adversarial loss is:
\begin{equation}
\mathcal{L_{\mathrm{IntersectGAN}}} = \ \sum_{1 \leq i \leq n}{\mathcal{L}_i}  + \lambda_{\mathrm{Intersect}} \mathcal{L}_{\mathrm{Intersect}},
\label{eq: full_loss}
\end{equation}
where $\lambda_{\mathrm{Intersect}}$ is a constant controlling the relative significance among different objectives. 
During the training phase, the optimization problem can be formulated as a minimax game where the generative model tries to minimize the objective while discriminators aim to maximize it, and the goal is to find appropriate parameters for the intersection generator.
This is given by:
\begin{equation}
G_{Y}^{*} = \arg \min_{G_{i}, G_Y} \max_{D_{i}} \mathcal{L_{\mathrm{IntersectGAN}}},\ \mathrm{where}\ 1 \le i \le n. 
\end{equation}
The optimization aims to find an appropriate set parameters where the intersection generator can produce most realistic samples.

The core idea of the proposed IntersectGAN model is reflected in two parts: 
\begin{enumerate}
    \item The existence of parallel generators $G_i$ is for boosting their adversaries $D_i$ to better distinguish real samples and fake samples;
    
    \item All the discriminators $D_i$ are the adversaries of $G_{Y}$, which allows the intersection generator to learn all the attributes presented in those given domains. 
\end{enumerate}

\subsection{Implementation}

\textbf{Revised Adversarial Loss Function} 
In order to stabilize the training outcomes and help models to converge better, we follow the strategy used in Wasserstein GAN \cite{WGAN} to define a new objective function by introducing the gradient penalty \cite{WGAN_GP}. 
Then the loss function defined in Equation (\ref{eqt: gan_loss}) can be re-formulated as: 
\begin{equation}
\begin{split}
\mathcal{L}_{\mathrm{GAN}}(X, G, D) & = \  \mathbb{E}_{x \sim p_X} \left[ D(x) \right]  - \ \mathbb{E}_{z \sim p_Z} \left[ D(G(z)) \right] \\
& - \ \lambda_{pg} \mathbb{E}_{\hat{x} \sim p_{\hat{x}}} \left[ {( \norm{\nabla_{\hat{x}}D(\hat{x})}_{2} - 1)}^{2}  \right], 
\end{split}
\end{equation}
where $\hat{x}$ is uniformly sampled from the linear space between a pair of real and generated images.

\textbf{Architecture}
Deep convolutional neural networks \cite{CDGAN} are used to implement IntersectGAN, which is constructed with $n+1$ structurally identical generator networks and $n$ discriminator networks.
When generating sample images with the size of $128 \times 128$, we fix the dimension of input noise to 32.
All the generators consist of one fully connected layer and seven deconvolutional layers. 
All the discriminators contain six convolutional layers and two fully connected layers to output a single logit indicating the probability whether the input sample is real.  
Batch normalization \cite{BatchNorm} and layer normalization \cite{LayerNorm} are utilized for generators and discriminators, respectively.
Note that the relationship among generators (discriminators) can be explored through weight sharing as explained in Section \ref{ssec:trio}.

\begin{figure}
\begin{center}
\includegraphics[width=83mm]{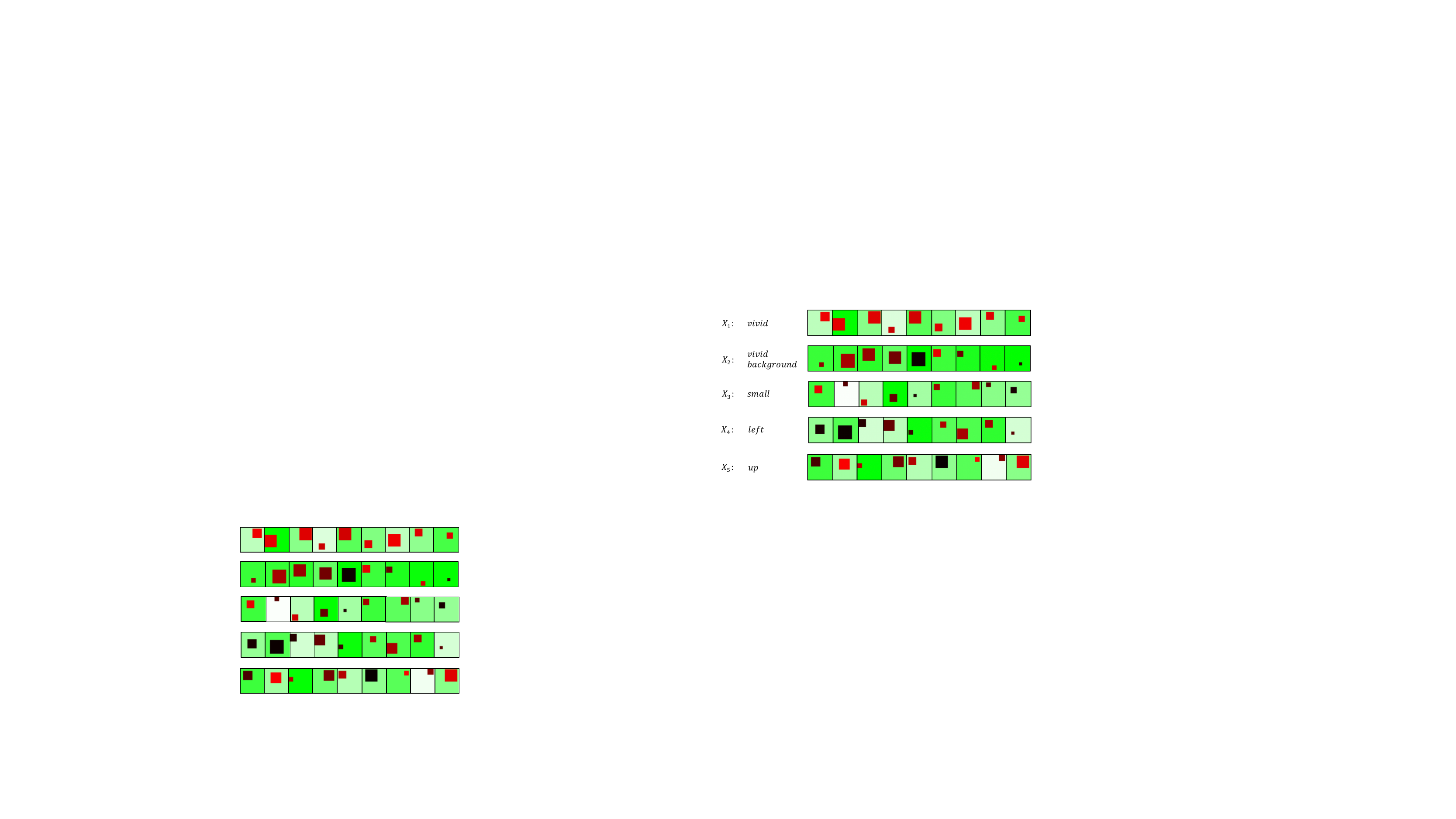} 
\end{center}
  \caption{Image samples from Colored Square dataset used in the experiments. Domains $X_1$, $X_2$, $X_3$, $X_4$ and $X_5$ are specified with attributes \textit{vivid}, \textit{vivid background}, \textit{small}, \textit{left} and \textit{up} respectively.
}
\label{fig:square_samples}
\end{figure}

\begin{figure*}
\begin{center}
\includegraphics[height=96mm]{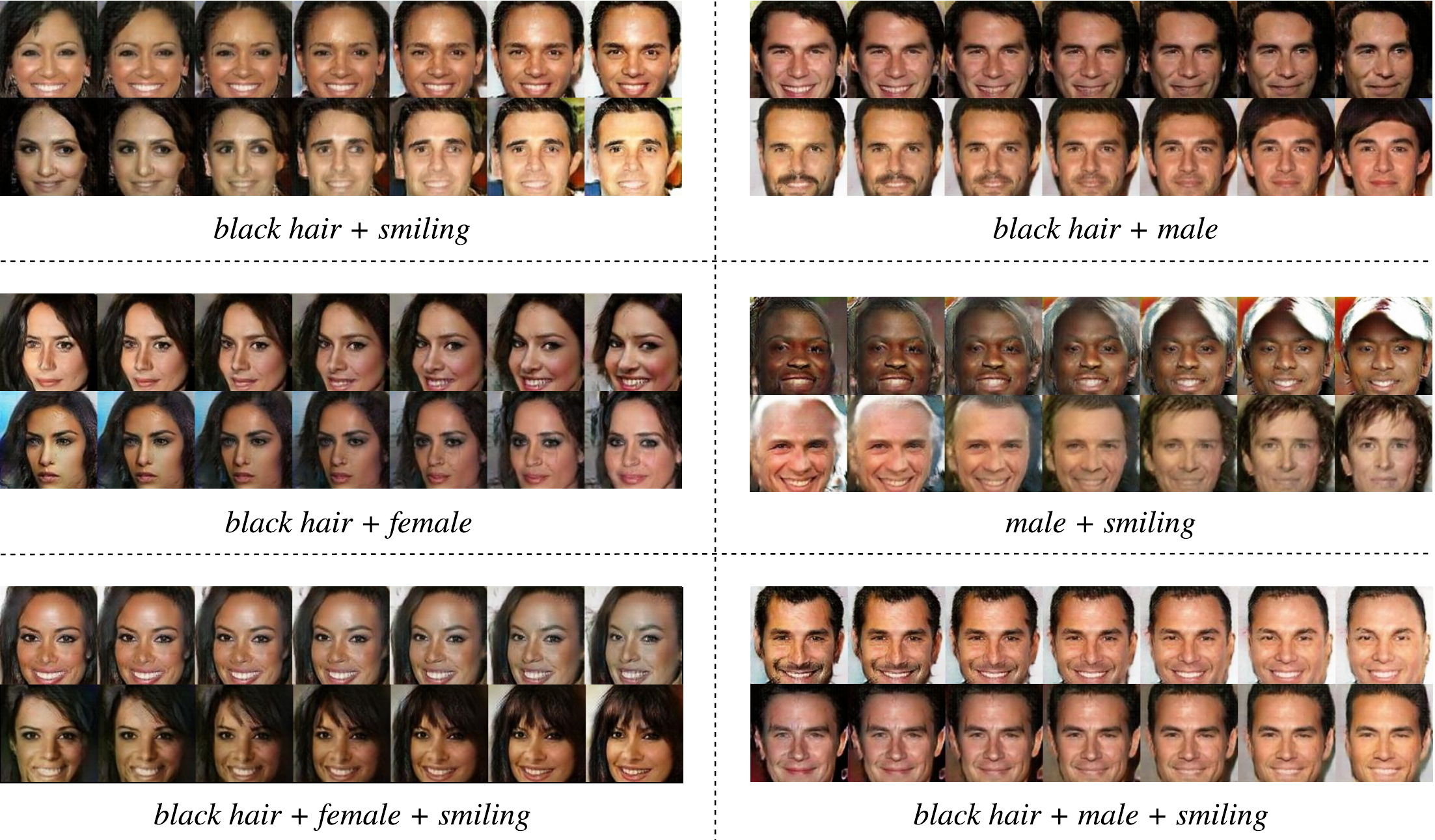}
\end{center} 
  \caption{Sample face images with two or three attributes that are generated by our proposed IntersectGAN. 
  } 
\label{fig:face_sample}
\end{figure*}

\textbf{Training} 
During the training phase, we apply random cropping and flipping to real input images in order to enhance sample variety. 
All the input vectors of random noise are sampled from the standard normal distribution.  
The batch size of input data is 32. 
We use Adam optimizer \cite{AdamOptimizer} to train the model with $\beta_1$ = 0.5 and $\beta_2$ = 0.999. 
In addition, the initial learning rate was set to 0.0002. 
We train the model for 20,000 iterations and apply linear decay on the learning rate in the second 10,000 iterations to stabilize the training phase.  
We assume all the attributes are equally important (i.e. $\alpha_i = 1.0$ for all $1 \le i \le n$) and set $\lambda_{\mathrm{Intersect}}$ and $ \lambda_{gp}$ to $1.0$ and $10.0$ respectively  in all the experiments.

\section{Experimental Results and Discussions}
\label{sec:exp}

\begin{figure}
\begin{center}
\includegraphics[width=80mm]{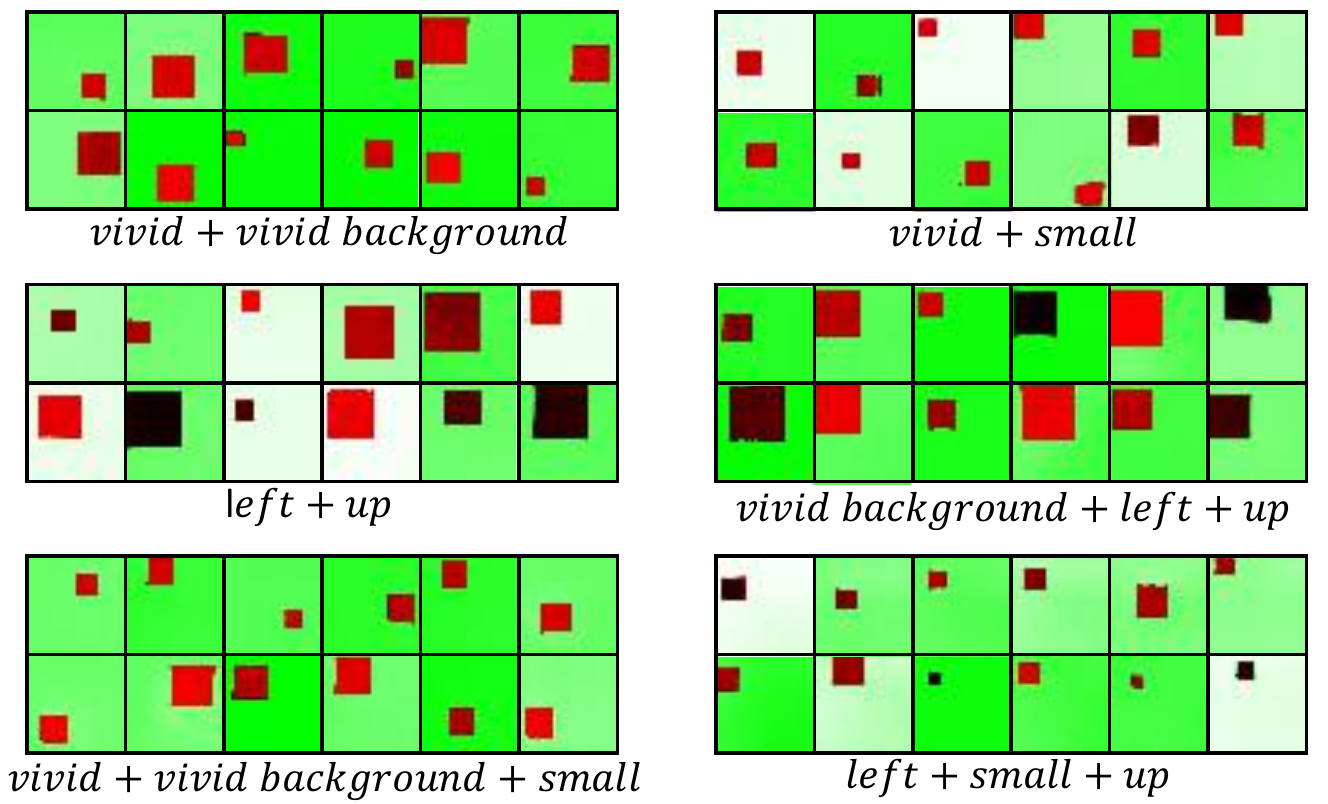} 
\end{center}
  \caption{Generated samples of images in different experiments with the Colored Squares dataset. 
}
\label{fig:square_results}
\end{figure}

In this section, we first present two experiments for qualitative evaluation: a preliminary experiment on generating colored squares, and a comprehensive experiment on generating face images with multiple attributes.
Then a set of experiments are introduced for quantitative evaluation against several baseline algorithms.

\subsection{Experiments} 

\textbf{Generating Multi-Attribute Squares} 
In the experiment, as illustrated in Figure \ref{fig:square_samples}, 5 specific visual attributes are utilized to describe the colored square images: \textit{vivid} (i.e. the color of a square is closer to pure red), \textit{vivid background} (i.e. the background color is closer to pure green), \textit{small}, \textit{left} and \textit{up}.
An image dataset of Colored Squares was purpose-built for validating the objective of IntersectGAN. 
It consists of 5 groups of images and each of them contain 2000 image samples of a colored square. 
Each group corresponds to one visual attribute and all the images in that group contain the a specific attribute and other attributes are randomly generated.
All the samples are generated randomly by our computer program and all the random factors (e.g. color) are sampled from uniform distributions. 
It is likely that there is no exact the same image that appears in two image groups.
In the experiment, the goal is to generate images possessing all the specific attributes after training our proposed IntersectGAN with the given training datasets. 
Sample results of each experiment can be found in Figure \ref{fig:square_results}, which indicates that IntersectGAN is able to generate image samples with multiple specific attributes.

\textbf{Generating Multi-Attribute Faces} We use CelebFaces Attri- butes   (CelebA) Dataset \cite{liu2015faceattributes} to train baseline models for generating faces images with multiple attributes. 
There are 202,599 annotated image samples of celebrity faces in the dataset.
According to the annotations, before training the model, we partition the dataset into different domains in terms of specified visual attributes.
As shown in Figure \ref{fig:face_sample}, our proposed IntersectGAN is able to produce high quality two-attribute and three-attribute face images.    
In each row, the samples generated are gradually changed from one face to the other by linearly changing the input noise vector.  
For example, each of the two-attribute samples {\it black hair} + {\it smiling} clearly present such two attributes, although the \textit{gender} attribute does not affect the results.

\begin{figure*}
\begin{center}
\includegraphics[width=170mm]{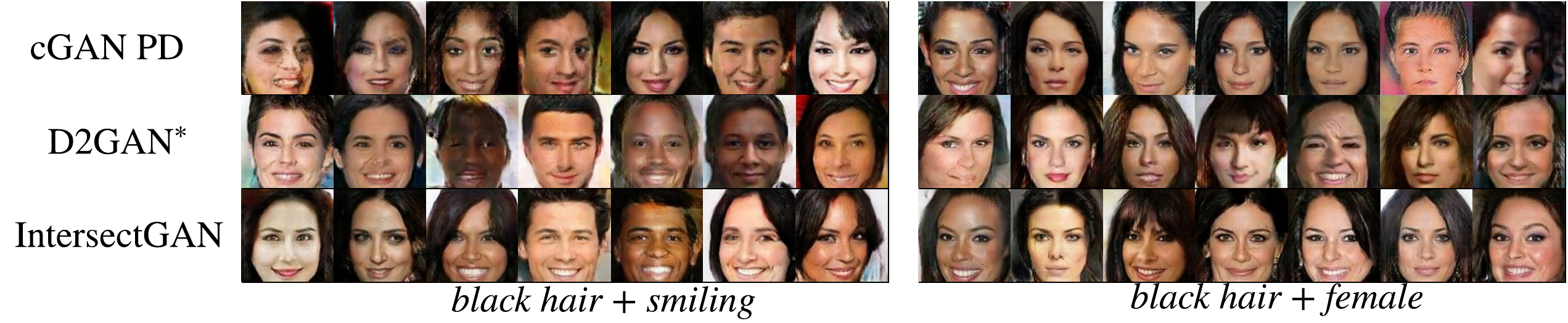}
\end{center}
\caption{
Sample multi-attribute face images generated by different GAN models trained on the CelebA images. The example on the left represents the combined attributes $black hair$ and $smiling$ and the example on the right represents the combined attributes $black hair$ and $female$. 
}
\label{fig:qualitative_comparison}
\end{figure*}

\subsection{Metrics}

\textbf{Fréchet Inception Distance (FID)} \cite{FID} As a quantitative measurement of GAN proposed recently, \textit{FID score} is used to estimate the generating quality of baseline models by measuring the Fréchet distance between generated samples and real samples. 
For each face generation experiment with certain combination of specified attributes, we collect ground truths from CelebA datasets according to the given annotations and face image samples generated by each baseline model that is supposed to simultaneously contain all the specified attributes.

\textbf{Perceptual Study} 
As evaluating image realism and identifying facial attributes can be subjective, we conduct two perceptual studies with Amazon Mechanical Turk (AMT) system.
To estimate the realism of the samples generated by each model, we obtain a \textit{Realism Score} for each model by using AMT to perform human based perceptual survey.
At first, 1000 groups of image samples were generated from different models (i.e., our proposed IntersectGAN and the baseline models) and each group contains one image sample generated by each model with the same noise input. 
In the survey, a Tucker was shown a number of image groups and asked to select the most realistic one from each group.
To ensure the quality of data collection from Turkers, we set additional qualifications to only allow workers whose Approval Rate is greater than $95\% $ to participate in the survey and also assign a few questions with apparent answers to validate their effort.
After all the image groups are examined, we obtain the Realism Score for each model by calculating the percentage of sample images generated by the model that Tuckers selected as the most realistic ones.

To further examine the presence or absence of specified attributes in the generated samples, we design another human based perceptual survey using AMT to obtain \textit{Attribute Score} for each model.
In the survey, a Turker was shown a number of generated image samples and for each given sample the Turker was asked to answer three multiple choice questions in regard to specific attributes including hair color, expression, and gender. 
For example, for gender attribute, the given choices are $\textit{Male}$, $\textit{Female}$ and $\textit{Cannot identify}$.
In order to reduce the bias caused by random selection, in the experiment, 300 image samples were generated by each model for examination.
After all the tasks are finished, we derive the Attribute Score for each model by calculating the percentage of image samples generated by the model that simultaneously possesses all the specified attributes.

\subsection{Baselines}

\textbf{Conditional GAN with Projection Discriminator (cGAN PD) } As one of the-state-of-art GAN-based architectures with additional supervision inputs, conditional GAN with projection discriminator \cite{cgan-pd} is used as one of the baseline model. $n$ labels (i.e., 0/1) are given to indicate whether an output image should respectively contain each attribute.
As a supervised learning method, in its implementation, the label of each image needs to be available, which means that it has stronger or stricter requirements than ours.

\textbf{GAN with Dual Discriminators (D2GAN*)} This model has the identical structure with the D2GAN model proposed by Nguyen et al. \cite{DualDisGAN}. Two discriminators in the original D2GAN learn from the same datasets. 
However, in the comparison experiments, they were trained from different image domains. 
When the number of specified attributes is more than 2, we increase the number of discriminators correspondingly.
The comparison with this model can help tell how much the quality of generated image samples would increase when a model contains multiple parallel generators. 
For the sake of convenience, we denote the modified model as D2GAN* in this paper.

\begin{table}
    \centering
    \begin{tabular}{|c|c|c|c|c|c|}
        \hline
        \textbf{Methods}  & \textbf{B+M} & \textbf{\textit{M+S}} & \textbf{\textit{F+S}} & \textbf{\textit{B+M+S}}  & \textbf{\textit{B+F+S}} \\
        \hline
        cGAN PD  & 84.6 & 93.8  & 69.9 & 94.2 & 73.5 \\
        \hline
        D2GAN*  & 90.0 & 96.9 & 64.9 & 97.0 & 81.6 \\
        \hline
        IntersectGAN & \textbf{77.9} & \textbf{79.9}  & \textbf{64.5} & \textbf{83.5} & \textbf{72.3}  \\
        \hline
    \end{tabular}
    \bigskip
    \caption{Comparison results in terms of FID for different attribute combinations of the experiments with CelebA datasets where B, F, M and S denote the attributes \textit{black hair}, \textit{female}, \textit{male} and \textit{smiling}, respectively.}
    \label{tab:fid_tab_CelebA} 
\end{table}

\subsection{Comparisons}
In each comparison experiment, we specify $n$  attributes and collect real samples of these attributes to build the training datasets $X_i$ for $1 \le i \le n$. 
The CelebA dataset is used in all the following comparison experiments. 
For fair comparison, we use an identical architecture to implement all the models and train them under the same configuration.

For qualitative comparison, as shown in Figure \ref{fig:qualitative_comparison}, the image samples generated by our proposed IntersectGAN are generally of highest visual quality, while those by other baselines are of lower visual quality (e.g., artifacts and unnatural skin color).  
In addition, the image samples generated by our IntersectGAN present clear attributes than those generated by other GANs, as IntersectGAN aims to {\it fool} all the discriminators corresponding to each attribute.

For quantitative comparison, we conduct experiments with 5 combined attributes, including \textit{black hair + male}, \textit{male + smiling}, \textit{female + smiling},  \textit{black hair + male + smiling}, and \textit{black hair + female + smiling}.
As illustrated in Table \ref{tab:fid_tab_CelebA}, our proposed IntersectGAN achieves lowest FID scores in all the comparative experiments. 
D2GAN* is inferior in all the comparisons  as well, which indicates the boosting effect of parallel generators.
Without any supervision labels, our proposed IntersectGAN performs even better than cGAN PD for the lower FID scores.
The comparison results indicate that our proposed IntersectGAN performs better for generating samples of higher quality and diversity than other baseline models. 

For perceptual studies, as shown in Table \ref{tab:amt_rs}, IntersectGAN achieves significantly higher Realism Score than other baselines models, which means that our IntersectGAN is able to generate more realistic image samples than other baseline models.
It is expected that the advantage will be more obvious when the number of attributes increases.

As shown in Table \ref{tab:amt_cs}, our IntersectGAN also achieves considerably high Attribute Scores in the experiments. 
Percentages of image samples possess all the specified attributes are given as a reference.
Our proposed IntersectGAN achieves highest Attribute Scores in all the comparisons which are much higher than the ground truth percentage. It demonstrates the advantage of our model that it is able to synthesize high quality images of multiple attributes without relying on real image samples possessing those attributes
As for cGAN PD and D2GAN*, in addition to the potentially poorer capability to enforce output images to contain both specified attributes, the low realism of their generating samples can also negatively influence the Attribute Score.

\begin{table}
    \centering
    \begin{tabular}{|c|c|c|c|c|c|}
        \hline
        \textbf{Methods}  & \textit{\textbf{B+M}} & \textbf{\textit{M+S}} & \textbf{\textit{F+S}} & \textbf{\textit{B+M+S}}  & \textbf{\textit{B+F+S}} \\
        \hline
        cGAN PD  & 31.7\% & 40.0\% & 22.6\% & 27.8\% & 21.8\% \\
        \hline
        D2GAN* & 16.2\% & 16.5\% & 23.9\% & 25.5\% & 32.5\% \\
        \hline
        IntersectGAN  & \textbf{52.1\%} & \textbf{43.5\%} & \textbf{53.5\%} & \textbf{46.7\%} & \textbf{45.7\%} \\
        \hline
    \end{tabular}
    \bigskip
    \caption{
    Comparison results in terms of Realism Score for different attribute combinations.
    }
    \label{tab:amt_rs} 
    \vspace{-0.1in}
\end{table}

\begin{table}
    \centering
    \begin{tabular}{|c|c|c|c|c|c|}
        \hline
        \textbf{Methods}  & \textbf{B+M} & \textbf{\textit{M+S}} & \textbf{\textit{F+S}} & \textbf{\textit{B+M+S}}  & \textbf{\textit{B+F+S}} \\
        \hline
        cGAN PD  & 80.6\% & 69.0\% & 83.0\%  & 41.6\% & 46.6\% \\
        \hline
        D2GAN*  & 52.6\% & 41\%  & 68.6\%  & 23.6\%  & 20.6\% \\
        \hline
        IntersectGAN  & \textbf{87.0\%} & \textbf{81.3\%} & \textbf{91.3\%} & \textbf{47.6\%}  & \textbf{53.6\%}  \\
        \hline
        \hdashline
        \hline
         Ground Truths  & 12.4\% & 16.7\% & 31.5\% & 5.1\% & 6.4\%  \\
        \hline
    \end{tabular}
    \bigskip
    \caption{Comparison results in terms of Attribute Score for different attribute combinations.}
    \label{tab:amt_cs} 
\end{table}

\begin{figure}[t]
\begin{center}
\includegraphics[height=36mm]{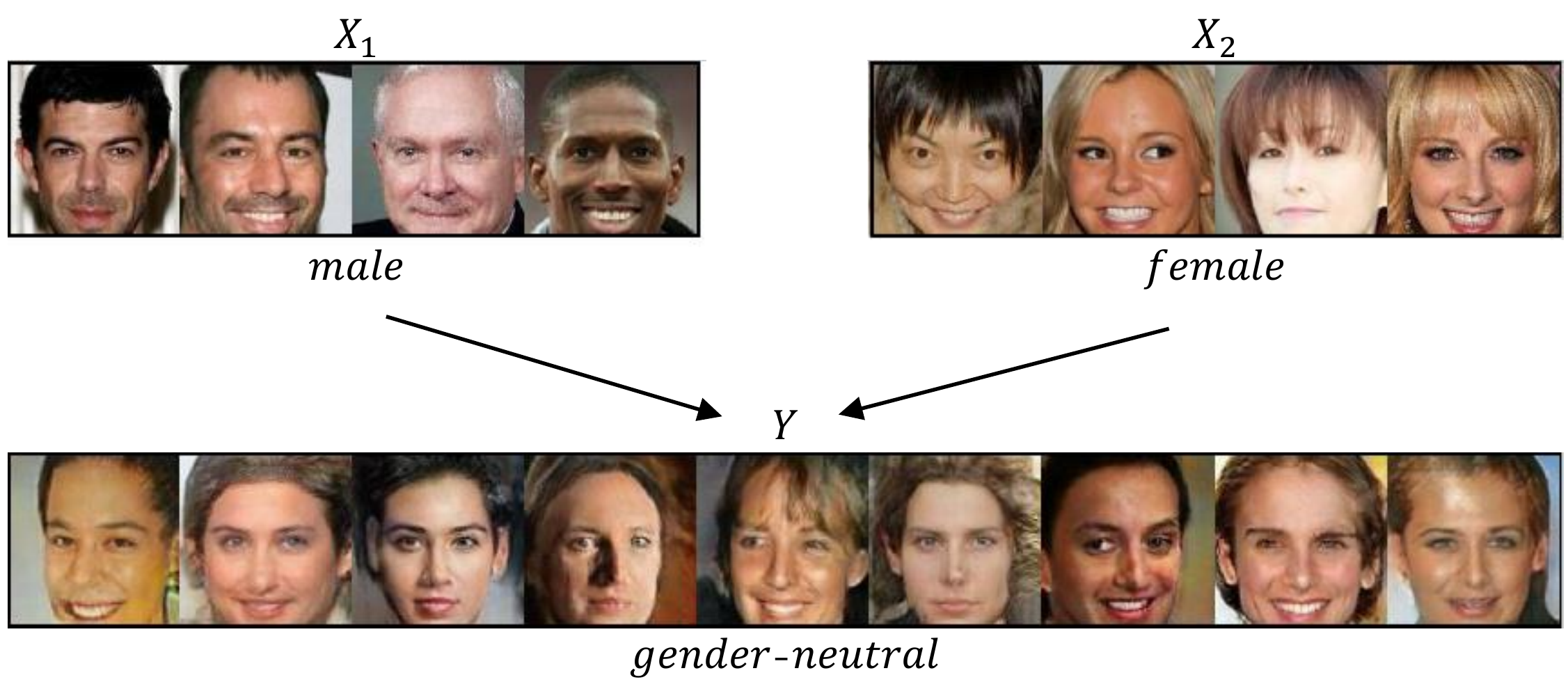} 
\end{center}
  \caption{Gender neutral image samples generated by our IntersectGAN from the \textit{male} and \textit{female} face domains. 
  }
\label{fig:male_female}
\end{figure}

\begin{figure}[t]
\begin{center}
\includegraphics[height=33mm]{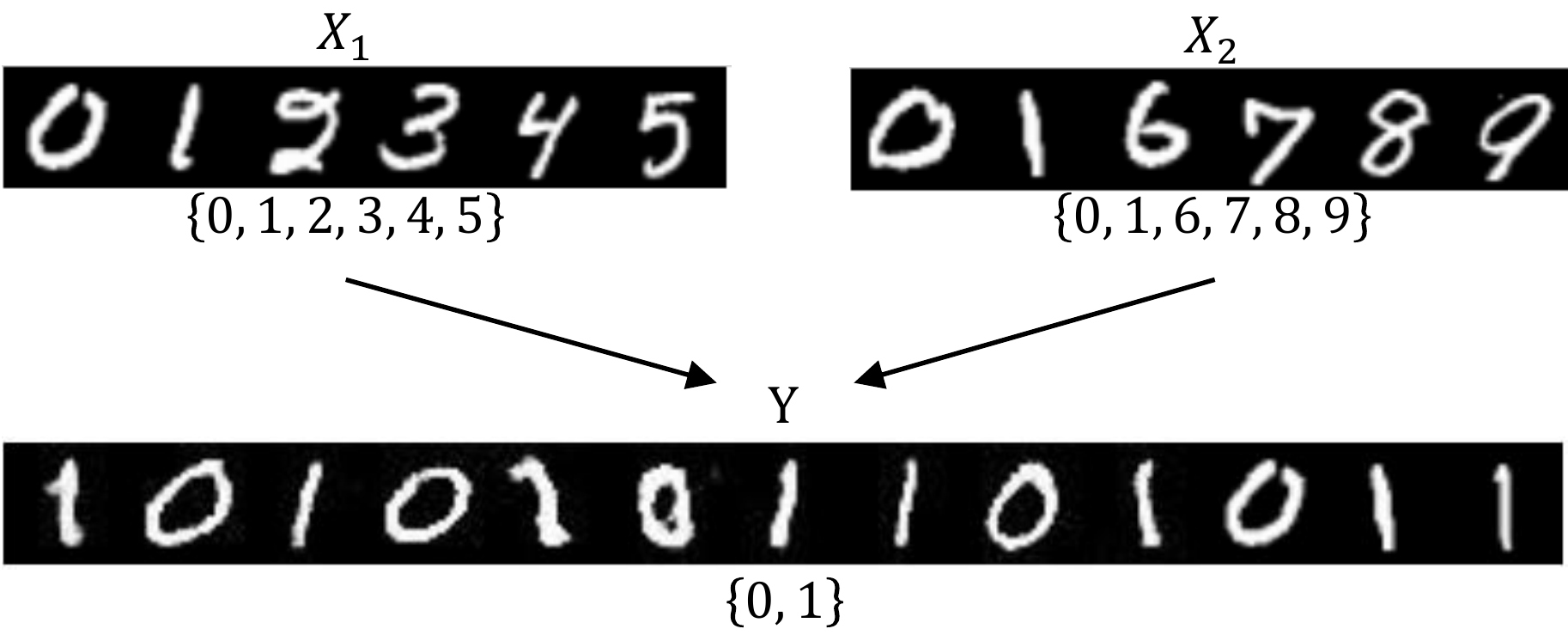} 
\end{center}
  \caption{
  Illustration of content-aware image domain intersection with the MNIST dataset. 
  Domain $X_1$ contains images of $0, 1, 2, 3, 4, 5$ and $X_2$ consists of $0, 1, 6, 7, 8, 9$. 
  The $X_1 \cap X_2$ row displays the generated samples of the intersected domain containing two digits $0$ and $1$ only. 
  }
\label{fig:mnist_example}
\end{figure}

\section{Other Applications}

In this section, we explore the generation capacity of IntersectGAN for three interesting applications: generating images of a blended attribute from two opposite attributes, generating content-aware domain intersected images, and generating trio images.

\subsection{Generating Images of Blended Attributes}

Semantic attribute indicates the presence or absence of certain characteristics in an image, which can be regarded as a binary value.
When these attributes are independent of each other, intersecting such attributes leads to the combination of these attributes.
However, when such attributes contradict to each other, such as gender, intersection would lead to a blending effect on such attributes, 
which is a unique outcome of our IntersectGAN, in comparison with cGAN and its variants.

In the experiment, we divide the CelebA dataset into two sets: {\it male} image set and {\it female} image set, according to the annotations provided. 
For IntersectGAN, domain $X_1$ and domain $X_2$ possess the attributes $male$ and $female$, respectively.
According to the intersection nature of IntersectGAN, it is expected that the generated image samples will possess both {\it male} and {\it female} attributes.

In other words, generated images of such two attributes would be gender-neutral. 
As shown in Figure \ref{fig:male_female}, it is not easy to tell the gender of the generated image samples in the bottom row.

\subsection{Generating Content-aware Domain Intersected Images} 
This task is to demonstrate that the proposed IntersectGAN is able to generate images with both content and attribute intersected from two domains. 
Assume we have two image domains $X_1$ and $X_2$ where $X_1$ contains images of content $c_{i_1}, c_{i_2}, ...$, and images in $X_2$ are of content $c_{j_1}, c_{j_2}, ...$. 
For both image domains, we denote the set of content items as $C_1$ and $C_2$, and the set of intersected content items as  $C_1 \cap C_2$.  

We train our model on the MNIST dataset \cite{mnist} for this task. As shown in Figure \ref{fig:mnist_example}, domain $X_1$ contains images of digits $0, 1, 2, 3, 4, 5$, domain $X_2$ contains images of digits $0, 1, 6, 7, 8, 9$, and the intersected domain $Y$ contains generated samples of 2 digits $0$ and $1$ only which is the intersection of two sets $ \left\{ 0, 1, 2, 3, 4, 5 \right\}$  and  $\left\{ 0, 1, 6, 7, 8, 9 \right\}$.

\begin{figure}[t]
\begin{center}
\includegraphics[height=34 mm]{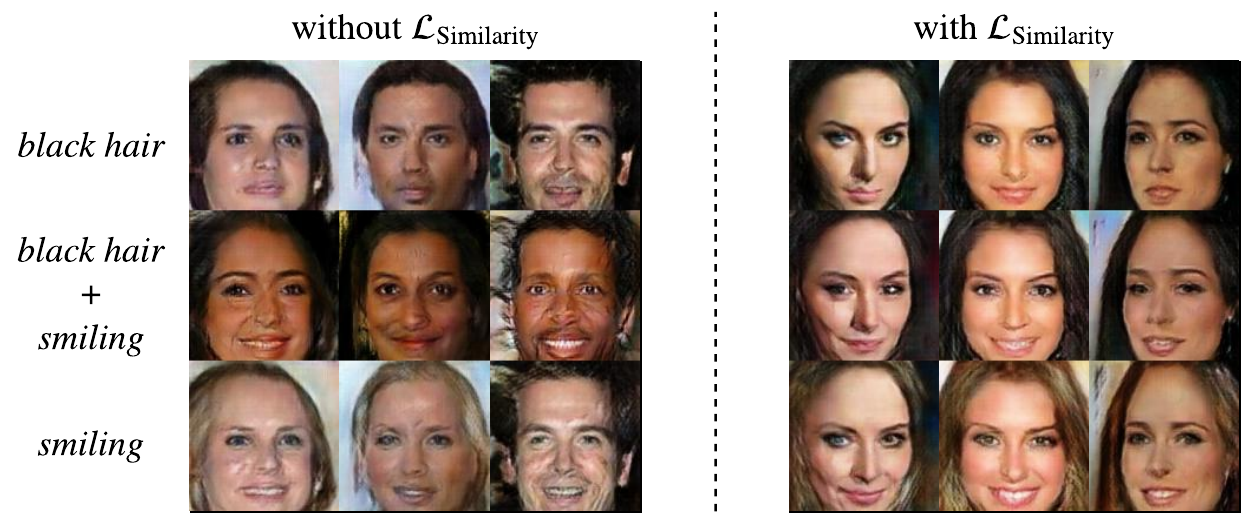} 
\end{center}
  \caption{The effect of similarity loss for generating trio image samples. The specified attributes in the comparative experiment are \textit{black hair} and \textit{smiling}.}
\label{fig:similarity_loss}
\end{figure}

\subsection{Generating Trio Images}
\label{ssec:trio}
\ \ \  \ To better visualize the objective of multi-attribute image generation by learning domain intersection, we change the network architecture of IntersectGAN to generate image samples in trio which are visually similar.
We follow the idea from CoGAN \cite{CoGAN} and share the  weights in the first several layers of three generators in our model. In addition, discriminators $D_1$ and $D_2$ share the weights except the first two layers. This architecture allows IntersectGAN to generate samples in pair. 
Then for each input noise $z$, we can generate samples in trio denoted as $G_1(z)$, $G_2(z)$ and $G_{Y}(z)$, respectively, where $G_1(z)$ and $G_2(z)$ are the samples produced by two parallel generators and the sample $G_{Y}(z)$ would contain the typical attributes of both domain $X_1$ and $X_2$. Ideally, all three generated samples are supposed to look visually similar in some ways.

In addition to the existing objective of IntersectGAN, to improve the correspondence of each trio of output samples, we introduce an additional loss for image generation, namely \textit{similarity loss} denoted as a L1 loss:
 \begin{equation}
 \begin{split}
     \mathcal{L}_{\mathrm{Similarity}} = & \ \mathbb{E}_{z \sim p_Z} \left[ \norm{G_1(z) - G_{Y}(z)}_{1}  \right] \\
  + & \  \mathbb{E}_{z \sim p_Z} \left[ \norm{G_2(z) - G_{Y}(z)}_{1}  \right].
 \end{split}
\end{equation}

Its objective is to enforce the similarity between $G_{Y}(z)$ (the output of the intersection generator) and $G_1(z)$ and $G_2(z)$ (the outputs of the two domain generators). 
A coefficient $\lambda_{\mathrm{Similarity}}$ is introduced to balance the importance of such similarity loss and we set it to $10$ in our experiments. 
Without $\mathcal{L}_{\mathrm{Similarity}}$, although the generated images $G_{Y}(z)$ contain both specified attributes (e.g., \textit{black hair} and \textit{smiling expression}), they may look far from the corresponding samples generated by the parallel generators on other aspects (e.g., skin color and image background), as shown in Figure \ref{fig:similarity_loss}.

\begin{figure}
\begin{center}
\includegraphics[height=30mm]{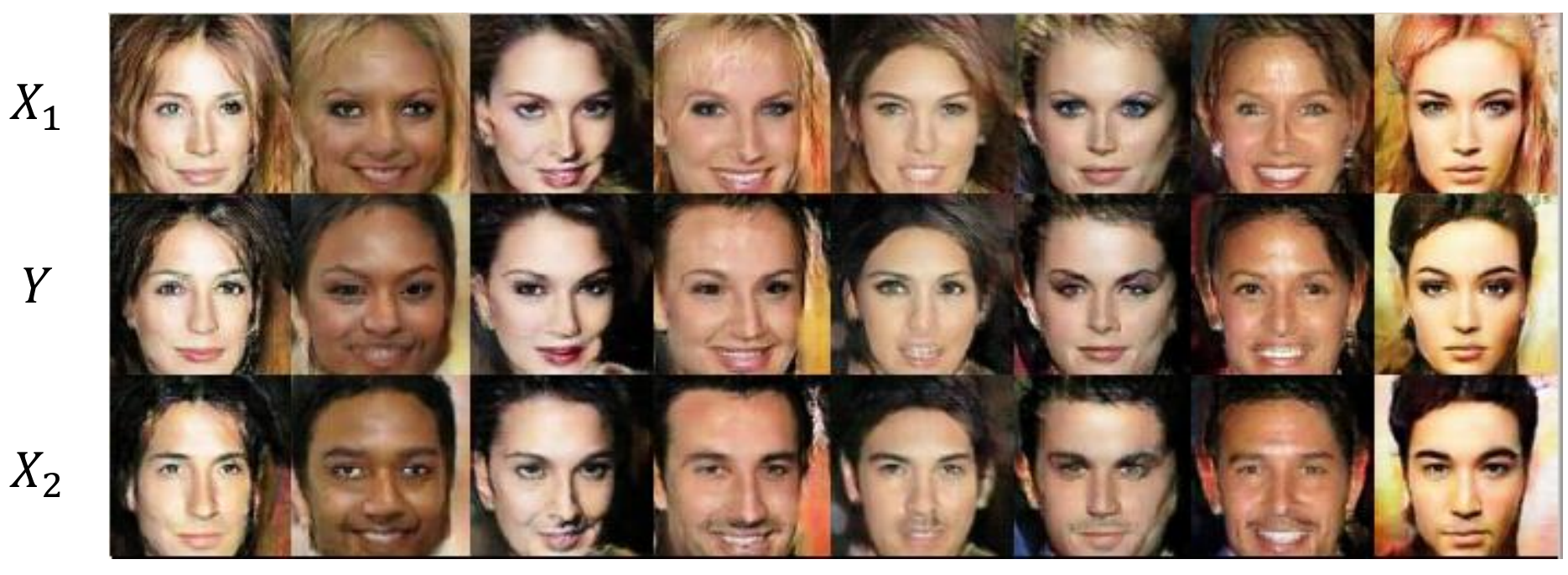}
\end{center}
  \caption{
  Illustration of trio-image samples. 
  From the top to bottom, image samples are generated by $G_1$, $G_Y$ and $G_2$. 
  The specified domain attributes are \textit{female} and \textit{black hair}. 
  }
\label{fig:samples_in_trio}
\end{figure}

As shown in Figure \ref{fig:samples_in_trio}, we generate several groups of trio-image samples with IntersectGAN.
We observed that each trio set appears visually similar while the image samples in domain $X_1 \cap X_2$ still contains the attributes intersected from domains $X_1$ and $X_2$.

\section{Conclusion}
In this paper we present a novel IntersectGAN framework to learn intersection of multiple image domains for generating image samples possessing multiple attributes without using real samples simultaneously possessing those attributes. 
To the best of our knowledge, this is the first GAN model generating multi-attribute images from noise input without introducing extra supervision labels.
Both qualitative and quantitative evaluations have demonstrated that our proposed IntersectGAN is able to produce high quality images possessing multiple attributes from separate image sets possessing each individual attribute, rather than from real image samples possessing multiple attributes which are generally expensive to collect.  
The generation capacity of the proposed IntersectGAN is further explored for three applications: generating images of a blended attribute, content-aware domain intersection based image generation, and trio-image generation.
In our future work, we will further extend this model to deal with more attributes with more scalable network architectures.

\section*{ACKNOWLEDGEMENTS} 
This research was partially supported by ARC (Australian Research Council) grant DP160103675.  Dong Xu is supported by ARC Future Fellowship FT180100116. 
We would like to thank the NVIDIA Corporation for GPU support through NVIDIA Hardware Grant.

\newpage

\bibliographystyle{ACM-Reference-Format}
\balance
\bibliography{sample-base}

\end{document}